\documentclass[journal]{IEEEtran}
\usepackage{graphicx} 
\usepackage{xcolor}
\usepackage{url}
\usepackage{glossaries}
\usepackage{booktabs}
\usepackage{multirow}
\usepackage[detect-none]{siunitx}
\usepackage{amsmath}
\usepackage{amsfonts}
\usepackage{amssymb}
\usepackage{pifont}
\usepackage[hidelinks]{hyperref}
\usepackage{orcidlink}

\newcommand{\cmark}{\ding{51}}%
\newcommand{\xmark}{\ding{55}}%


\DeclareSIUnit{\nothing}{\relax}
\DeclareSIUnit{\frame}{frame}

\usepackage{fancyhdr}
\title{Multi-modal On-Device Learning for Monocular Depth Estimation on Ultra-low-power MCUs}

\author{
Davide~Nadalini\orcidlink{0000-0002-8003-7633}, 
Manuele~Rusci\orcidlink{0000-0001-7458-4019}, 
Elia~Cereda\orcidlink{0000-0001-8888-3538},~\IEEEmembership{Student Member,~IEEE,} 
Luca~Benini\orcidlink{0000-0001-8068-3806},~\IEEEmembership{Fellow,~IEEE,} 
Francesco~Conti\orcidlink{0000-0002-7924-933X},~\IEEEmembership{Senior~Member,~IEEE,} 
and Daniele~Palossi\orcidlink{0000-0003-4487-0836}
\thanks{This work has been partially funded by the Swiss National Science Foundation (SNSF) RoboMix2 project (g.a. 10004854) and by the ChipsJU project ISOLDE (g.a. 101112274).}
\thanks{Davide Nadalini is with the Department of Electrical, Electronic, and Information Engineering (DEI), University of Bologna, 40126 Bologna, Italy, and also with the Department of Control and Computer Engineering (DAUIN), Politecnico di Torino, 10129 Torino, Italy (e-mail: d.nadalini@unibo.it).}
\thanks{Manuele Rusci is with Department of Electrical Engineering (ESAT) at KU Leuven, 3000 Leuven, Belgium. }
\thanks{Elia Cereda is with Dalle Molle Institute for
Artificial Intelligence (IDSIA), USI-SUPSI, 6962 Lugano, Switzerland.}
\thanks{Luca Benini is with the Department of Information Technology and
Electrical Engineering (D-ITET), ETH Zurich, 8092 Zurich, Switzerland, and also with the Department of Electrical, Electronic, and Information Engineering (DEI), University of Bologna, 40126 Bologna, Italy.}
\thanks{Francesco Conti is with the Department of Electrical, Electronic, and Information Engineering (DEI), University of Bologna, 40126 Bologna, Italy.}
\thanks{Daniele Palossi is with Dalle Molle Institute for Artificial Intelligence
(IDSIA), USI-SUPSI, 6962 Lugano, Switzerland, and also with the Integrated
Systems Laboratory, ETH Zurich, 8092 Zurich, Switzerland.}
}

\newacronym{IoT}{IoT}{Internet-of-Things}
\newacronym{MCU}{MCUs}{MicroController Units}
\newacronym{ULP}{ULP}{Ultra-low-power}

\DeclareSIUnit{\pixel}{px}

\begin{document}

\maketitle

\begin{abstract}

    Monocular depth estimation (MDE) plays a crucial role in enabling spatially-aware applications in Ultra-low-power (ULP) Internet-of-Things (IoT) platforms. 
    However, the limited number of parameters of Deep Neural Networks for the MDE task, designed for IoT nodes, results in severe accuracy drops when the sensor data observed in the field shifts significantly from the training dataset.
    To address this domain shift problem, we present a multi-modal On-Device Learning (ODL) technique, deployed on an IoT device integrating a Greenwaves GAP9 MicroController Unit (MCU), a 80~mW monocular camera and a $8 \times 8$ pixel depth sensor, consuming $\sim$\SI{300}{\milli\watt}.
    In its normal operation, this setup feeds a tiny \SI{107}{\kilo\nothing}-parameter $\mu$PyD-Net model with monocular images for inference.
    The depth sensor, usually deactivated to minimize energy consumption, is only activated alongside the camera to collect pseudo-labels when the system is placed in a new environment. 
    Then, the fine-tuning task is performed entirely on the MCU, using the new data.
    To optimize our backpropagation-based on-device training, we introduce a novel memory-driven sparse update scheme, which minimizes the fine-tuning memory to 1.2 MB, 2.2$\times$ less than a full update, while preserving accuracy (i.e., only 2\% and 1.5\% drops on the KITTI and NYUv2 datasets).
    Our in-field tests demonstrate, for the first time, that ODL for MDE can be performed in 17.8~minutes on the IoT node, reducing the root mean squared error from 4.9 to \SI{0.6}{\meter} with only 3~k self-labeled samples, collected in a real-life deployment scenario.

\end{abstract}


\section{Introduction}\label{sec:introduction}



A new generation of spatially aware miniaturized Internet-of-Things (IoT) devices is rapidly taking shape, including palm-sized autonomous quadrotors (nano-drones)~\cite{cereda2024training}, mixed-reality eyewear~\cite{novac2022uca}, and smart cameras~\cite{lee2025hypercam}. 
These novel devices feature Ultra-low-power (ULP) processing units that typically run Artificial Intelligence (AI) algorithms to analyze sensor data in real-time.
A key algorithmic task is monocular depth estimation (MDE)~\cite{ming2021deep}, which takes data from an inexpensive single-camera sensor and extracts depth map information, crucial for a large variety of applications, ranging from human pose estimation~\cite{srivastav2019human}, people counting~\cite{sun2019benchmark}, 3D scene understanding~\cite{chen2019towards}, vehicle detection~\cite{shen2020joint}, and obstacle avoidance \cite{yang2019fast}. 
MDE is critical for low-power and low-cost IoT devices, as it avoids the need for bulky stereo systems or costly and power-hungry long-range depth sensors, such as LIDARs~\cite{lee2020accuracy}.

Recent MDE approaches show compelling results by employing Deep Neural Network (DNN) models~\cite{eigen2014depth, godard2019digging, yang2025depth}.
Such models typically run on high-end computing devices, ranging from GPU-equipped desktop computers to smartphones, which can afford to execute large DNNs featuring billions of parameters~\cite{poggi2018towards, she2024evitbins,yang2025depth}.
Conversely, ULP MicroController Units (MCUs), with a power envelope of a few 100s of \SI{}{\milli\watt}, limited on-chip memory (i.e., a few \SI{}{\mega\byte} at most) and a computational power up to a few hundred GOPS/s, make the deployment of State-of-the-Art (SoA) models featuring 0.5-1 billion parameters~\cite{hu2024metric3d, yang2025depth, zhao2022monovit} prohibitive.
However, Peluso et al.~\cite{peluso2021monocular} have recently introduced a lightweight DNN model for MDE, called $\mu$PyD-Net, which requires only 0.1 M parameters (1000$\times$ less than SoA) and demonstrated its deployment on an MCU for IoT nodes. 
Unfortunately, this class of tiny MDE DNNs achieves acceptable accuracies only on test data distributions close to the training dataset, which is generally collected and then labeled using precise, yet bulky equipment.
As we also report in our study, when exposed to environment data diverse from the training set, these tiny DNNs are prone to large accuracy drops vs. SoA models because of the reduced capacity and the low generalization ability.

\begin{table*}[t]
\centering
\caption{Comparison of DNN-based methods for MDE. The scores refer the evaluation on the KITTI dataset}
\label{tab:relatedwork}
\resizebox{\textwidth}{!}{
\def\arraystretch{1.1}
\begin{tabular}{lcclccccc}
\toprule
\textbf{Model} &
\textbf{Year} &
\textbf{Params [M]} &
\textbf{Computation device} &
\textbf{Device class} &
\textbf{Power [W]} &
\textbf{On-device learning} &
\textbf{Image [px]} &
\textbf{$\delta_1$ [\%]} \\
\midrule
MonoViT~\cite{zhao2022monovit}              & 2022 & 10.0  & Nvidia RTX 3090    & Desktop GPU  & 350 & \xmark  & 640$\times$192 & 89.5 \\
Depth Anything V2~\cite{yang2025depth}      & 2025 & 25.0  & Nvidia V100        & Desktop GPU  & 250 & \xmark & 518$\times$518 & 97.3 \\
Metric3D v2, ViT-S~\cite{hu2024metric3d}    & 2024 & 22.0  & Nvidia A100        & Desktop GPU  & 250 & \xmark & 1064$\times$616 & 95.0 \\
Metric3D v2, ViT-L~\cite{hu2024metric3d}    & 2024 & 306.0 & Nvidia A100        & Desktop GPU  & 250 & \xmark & 1064$\times$616 & \textbf{98.5} \\
\midrule
PyD-Net~\cite{poggi2018towards}             & 2018 & 1.9 & Raspberry Pi 3       & Embedded GPU  &  5  & \xmark     & 512$\times$256 & 78.9 \\
EdgeNet~\cite{liu2022lightweight}           & 2022 & 1.7 & Nvidia Jetson Nano   & Embedded GPU  & 10  & \xmark     & 912$\times$228 & 85.4 \\
Liu \textit{et al.}~\cite{liu2022real}      & 2022 & 2.0 & Raspberry Pi 3       & Embedded GPU  &  5  & \xmark     & 512$\times$256 & 81.5 \\
LightDepthNet~\cite{liu2023lightdepthnet}   & 2023 & 1.3 & Nvidia Jetson Nano   & Embedded GPU  & 10  & \xmark     & 640$\times$192 & \textbf{86.0} \\
RT-Monodepth~\cite{feng2024real}            & 2024 & 1.2 & Nvidia Jetson Nano   & Embedded GPU  & 10  & \xmark     & 640$\times$192 & 84.0 \\
\midrule
$\mu$PyD-Net~\cite{peluso2021monocular}     & 2021 & 0.1 & STM32F7              & MCU  & 0.4 & \xmark & 48$\times$48 & 73.5 \\
\textbf{Ours}                               & 2025 & 0.1 & GWT GAP9             & MCU  & 0.1 & \cmark & 48$\times$48 & \textbf{80.8} \\
\bottomrule
\end{tabular}
}
\end{table*}

To tackle this fundamental domain shift problem, this work proposes a novel On-Device Learning (ODL) system-level method for tiny MDE models, deployed on IoT sensor nodes powered by MCUs. 
The ODL paradigm has recently been proposed to incrementally fine-tune DNNs directly on-device, based on locally sensed data and without relying on external compute resources~\cite{zhu2024device}.
In our solution, we leverage a multi-modal sensing frontend to address the absence of labeled data in the field. 
More in detail, we couple an always-on low-power camera consuming 80 mW with a power-hungry ($\sim$\SI{300}{\milli\watt}) depth sensor, which is activated only to produce the depth ground truth. 
With respect to high-end LIDAR or depth cameras, our depth sensor provides low-resolution depth maps ($8 \times 8$~\si{\pixel}) and can be integrated within an IoT node, thanks to its small form factor ($6.4 \times 3$~mm). 
Still, its power consumption is $>3\times$ higher than a typical ULP MCU. 

When pseudo-labeled data is collected without human intervention by matching image and depth information, the deployed MDE DNN undergoes on-device fine-tuning.
The learning task relies on backpropagation and is executed on a Greenwaves Technologies GAP9 multi-core MCU thanks to half-precision floating-point Single-Instruction Multiple-Data (SIMD) hardware.
We propose a new sparse update logic to update the DNN parameters of only a subset of layers based on a memory-driven criterion to reduce the high cost of on-device training. 
Our approach minimizes the memory requirements of the backpropagation algorithm while preserving fine-tuning accuracy vs. full-model fine-tuning.

We summarize our contributions as follows:
\begin{itemize}
    \item We introduce a multi-modal ODL technique to mitigate domain shift in tiny DNNs deployed for MDE on MCU-powered IoT nodes. 
    \item We propose a novel memory-driven sparse update logic to minimize memory usage and training time for on-device backpropagation in tiny U-Net-like models~\cite{azad2024medical}. 
    \item We extensively evaluate our ODL technique on public datasets to analyze how the two primary ODL challenges - the limited number of samples and the noisy labels - affect the fine-tuning accuracy. 
    \item We collect and publicly release IDSIA-$\mu$MDE, a custom dataset for MDE on IoT devices, which we use to evaluate our ODL method in a real-world scenario.
\end{itemize}

We field-test our technique on a GAP9Shield IoT node~\cite{muller2024gap9shield}, which features a GAP9 MCU, a QVGA Omnivision OV5647 mono-camera, and an ST Microelectronics VL53L5CX depth sensor, which produces $8 \times 8$~\si{\pixel} noisy depth labels.
When evaluated on the public KITTI and NYUv2 datasets, a $\mu$PyD-Net model with 107 k-parameters achieves $\delta_1$ accuracies of, respectively, 61.0\% and 47.0\% after ODL with our sparse-update. 
These scores are +43.2\% and +43.4\% higher than the present baseline solutions, where the DNN is trained off-device before the deployment on the MCU~\cite{peluso2021monocular}.
On the IDSIA-$\mu$MDE dataset, we show that the sparse ODL process can be performed with a single epoch of training in a memory budget of \SI{1.2}{\mega\byte}, taking only 17.8~minutes and 204.9~J for data collection and fine-tuning.
In this setting, a $\mu$PyD-Net adapts to the new environment using 3~k new images, and the RMSE notably reduces from 4.9~m to 0.6~m.
To the best of our knowledge, this is the first work to demonstrate the feasibility of ODL in MDE on IoT nodes, leveraging only low-resolution labels acquired with on-board sensors. 
%
From an application viewpoint, our solution enables the replacement of bulky and power-hungry depth or stereo cameras in IoT systems with low-cost monocular cameras, complemented by DNN-based processing that dynamically adapts to the target environment. 
This approach potentially makes embedded applications such as human pose estimation~\cite{srivastav2019human}, obstacle avoidance~\cite{zheng2024monocular}, and people counting~\cite{burbano20153d} more autonomous, cost- and energy-efficient.
We release the code of our experiments
\footnote{Link to our code: \url{https://github.com/idsia-robotics/ultralow-power-monocular-depth-ondevice-learning}}.


\section{Related Work}\label{sec:realted_work}

\subsection{Monocular Depth Estimation on Embedded Devices}

Several recent works propose memory-efficient MDE models that target 5-\SI{10}{\watt} edge platforms~\cite{liu2022lightweight,poggi2018towards,liu2022real,feng2024real,liu2023lightdepthnet,peluso2021monocular}. 
As reported in Table~\ref{tab:relatedwork}, these models all include a few million parameters, a $10\times$ reduction from the unconstrained highest-scoring DNNs~\cite{hu2024metric3d,yang2025depth,zhao2022monovit} benchmarked on popular public datasets such as KITTI~\cite{Geiger2013IJRR} or NYUv2~\cite{silberman2012indoor}.  
Among the proposed embedded solutions, EdgeNet~\cite{liu2022lightweight} introduces an encoder-decoder architecture with upsampling modules, reducing the model size down to \SI{1.7}{\mega\nothing} parameters with pruning and enabling \SI{55}{\milli\second}  inference on an NVIDIA Jetson Nano via custom GPU scheduling.
PyD-Net~\cite{poggi2018towards} employs a \SI{1.9}{\mega\nothing}-parameter pyramidal multi-scale architecture trained with an unsupervised stereo image reconstruction loss, that achieves \SI{2}{\hertz} inference on a Raspberry Pi~3.

In contrast, we are interested in IoT nodes that consume less than \SI{500}{\milli\watt} and can only afford to employ ULP MCUs.
Peluso et al.~\cite{peluso2021monocular} introduce $\mu$PyD-Net, the first and, to this date, only  MDE solution for this class of devices. 
Its 107~k-parameter encoder-decoder architecture can run on an STM32F7 MCU (i.e., a \SI{400}{\milli\watt} power envelope) to predict $48 \times 48$~\si{\pixel} depth maps with a latency of \SI{651}{\milli\second}.

Unlike the unconstrained SoA approaches that can work in any domain thanks to the high generalization ability~\cite{hu2024metric3d,yang2025depth}, $\mu$PyD-Net and the other edge memory-efficient works are still trained from scratch and tested on a specific domain (e.g., KITTI).
To overcome this limitation, this work proposes the first end-to-end system method for fine-tuning tiny MDE models on-device, using data acquired from the deployment environment and the associated ground truths captured by a depth sensor.
On the $\mu$PyD-Net architecture, our method improves the accuracy ($\delta_1$) of a model, pre-trained on a generic dataset, by +63\%, using just \SI{2.3}{\kilo\nothing} images from the target environment.
This corresponds to an accuracy of up to 80.8\%, comparable to models $20\times$ its size~\cite{poggi2018towards,liu2022real}.
 
We emphasize that, following Depth Anything V2~\cite{yang2025depth}, we adopt a fully-supervised training strategy, proven to achieve SoA accuracy on benchmarks such as KITTI. 
Specifically, we pre-train our $\mu$PyD-Net on synthetic data with perfect ground-truth labels and fine-tune it using $8 \times 8$ pseudo-labels from the on-board low-resolution depth sensor. 
Nevertheless, since our ODL technique is agnostic to the pre-training paradigm, it can be readily applied to models that have been initially pre-trained using self-supervised learning approaches~\cite{oquab2023dinov2}.

\subsection{Sparse On-Device Learning}

The growing demand for fine-tuning DNNs on-device has recently increased interest in novel hardware and software optimizations for ODL~\cite{zhu2024device}.
In the context of ULP MCUs, a few hardware-optimized ODL frameworks can now produce the source code for the backpropagation-based training task~\cite{nadalini2023reduced,wulfert2024aifes}. 
As a notable example, AIfES~\cite{wulfert2024aifes} uses the optimized software primitives from the ARM CMSIS-DSP library to accelerate the training kernels on ARM Cortex M CPUs.  
PULP-TrainLib~\cite{nadalini2023reduced} targets instead RISC-V multi-core platforms and leverages parallel processing and half-precision floating-point Single-Instruction-Multiple-Data (SIMD) instructions to speed-up the backpropagation task. 
When benchmarked on a 10-core RISC-V MCU, the PULP-TrainLib primitives can reach a top efficiency of 6.62~Multiply and Accumulate operations per clock cycle (MAC/clk) for convolutional layers. 
In this work, we build on PULP-TrainLib to bring, for the first time (to the best of our knowledge), an ODL pipeline for MDE on an MCU device. 

An orthogonal optimization axis concerns the usage of sparse update strategies that train (on-device) only a subset of parameters, reducing memory and computation costs vs. retraining the full model weights. 
In this context, Ren et al.~\cite{ren2021tinyol} propose to add a single trainable layer on top of a frozen and quantized model.
This fine-tuning task can run in milliseconds on an ARM Cortex-equipped Arduino board. 
TinyTL~\cite{cai2020tinytl} introduces an extra trainable layer with a low number of parameters in parallel to the frozen convolutional blocks, achieving 12.9$\times$ memory savings with respect to training all parameters of the original model. 
Paissan et al.~\cite{paissan2024structured} uses a structured sparse update scheme to reduce the gradient and the activation storage requirements in the backward step. 
This approach brings a 1.4~M-parameter CNN to learn two unseen classes in $< 20$ minutes when accelerated with PULP-TrainLib, reducing the computational and memory costs by 30\% with respect to a full update.
Tiny Training Engine~\cite{lin2022device} (TTE) extends the sparse update approach to quantized models already deployed on MCUs. 
By incrementally training only a subset of weights and layers, which are identified by computing an importance score with respect to a given benchmark, TTE reduces by up to 21$\times$ the training memory with respect to a full update.  

In contrast to the previous works that focus on DNN tailored for classification tasks, we develop a new sparse update logic for a tiny U-Net-like model ($\mu$PyD-Net).
We propose to retrain only the first block of the decoder, which requires $2.1\times$ and $1.3\times$ less memory than, respectively, fine-tuning all parameters of the last decoder block.
Thanks to sparse update and hardware-optimized ODL kernels, we enable the fine-tuning of $\mu$PyD-Net with a throughput of 3.57~samples/s, consuming only 41.3~mJ/sample. 

\subsection{On-device Learning with Unlabeled Data}

State-of-the-art MDE models are generally pretrained on large image datasets and then fine-tuned to the target downstream task using human-annotated data before deployment on the target platforms~\cite{yang2024depthv1,bhat2023zoedepth}. 
In contrast to this setting that requires task-specific labeled data to be available already at design time, post-deployment fine-tuning with unlabeled data has recently gained attention~\cite{sunaga2024sequential,lee2020learning}. 
Matsutani et al.~\cite{matsutani2024tiny} suggest a custom design of a low-power hardware core for ODL in a human activity classification task.
To deal with the lack of annotations during in-field operation, the system assigns pseudo-labels to new sampled data.
Specifically, if the prediction confidence returned by the inference task falls below a threshold, the new data is transmitted to a high-power external mobile device. This device invokes a teacher DNN model to generate high-quality pseudo-labels.
The on-device fine-tuning method for voice command classification sensors proposed by Rusci et al.~\cite{rusci2024self} assigns instead pseudo-labels to new samples based on a similarity score w.r.t. a few human-labeled samples stored in the IoT node’s memory.
Differently from these works, our work addresses a more challenging regression task, i.e. labels do not belong to a discrete set of values, and we rely on an additional sensor for generating the fine-tuning labels. 

As a notable application of ODL to regression problems, Boldo et al.~\cite{boldo2024domain} propose a teacher-student approach for human pose estimation and object detection. 
First, the non-real time teacher model is used to generate pseudo labels. 
Then, the student, which is originally distilled from the teacher, is fine-tuned on-device (NVIDIA Jetson Xavier NX) using the new set of labeled data. 
Cereda et al.~\cite{cereda2024training} address the domain shift observed for visual human pose estimation tasks after deployment on MCU-powered nano-drones. 
The proposed method leverages the temporal correlation of the images in a video to propagate human pose labels to successive frames based on the measured drone's motion. 
Thanks to the new labeled data, the parameters of the on-board visual DNN model are updated without relying on external compute resources. 

In contrast with these works, our solution adopts an auxiliary sensor to generate the ground truths, as label propagation through time cannot be applied to our regression problem. 
Thanks to our multi-modal sensor solution, new labels are produced with a throughput of 15~Hz on the MCU, 2.5$\times$ faster than the teacher model in~\cite{boldo2024domain}.
Overall, our work demonstrates, for the first time, an ODL approach to a regression MDE problem that can run entirely on a ULP IoT node.

\section{Background}\label{sec:hardware}

\begin{figure}[t]
\centering
\includegraphics[width=1\linewidth]{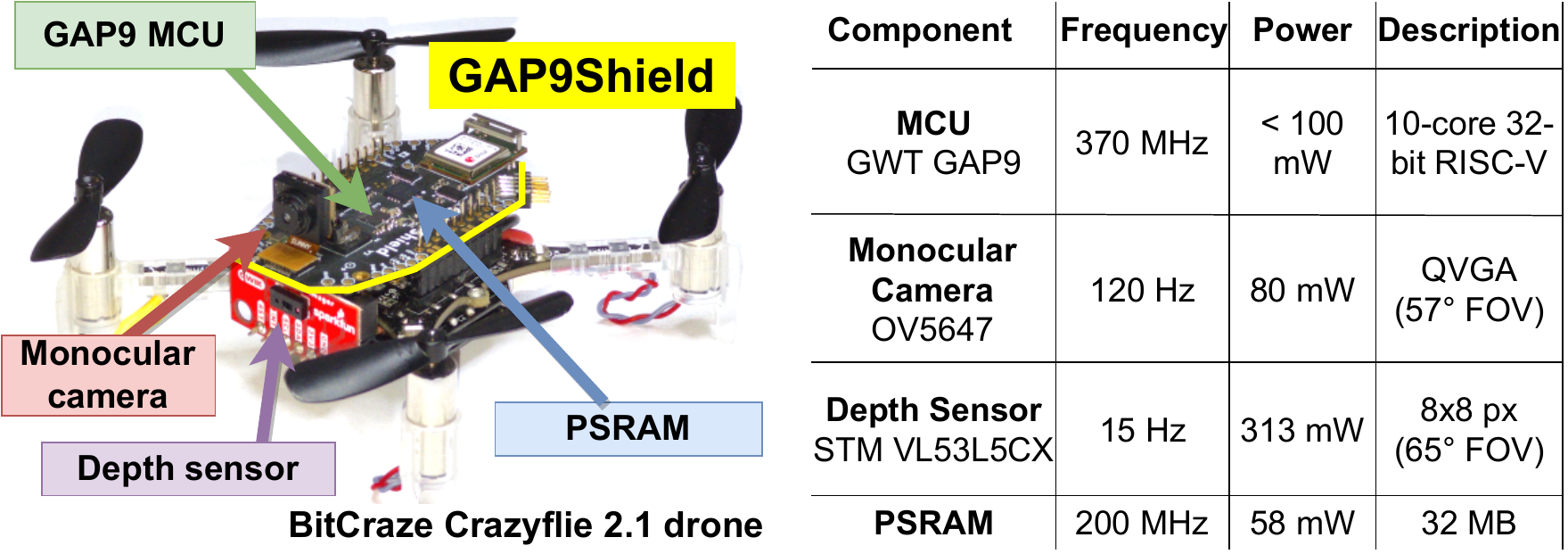}
\caption{\textit{GAP9Shield} IoT node \cite{muller2024gap9shield} mounted on a Crazyflie 2.1 (left) and the characteristics of the main components  (table in the right).}
\label{fig:iotnode}
\end{figure}

We consider the \textit{GAP9Shield}~\cite{muller2024gap9shield} as the reference IoT node platform to demonstrate MDE on-device learning. 
This state-of-the-art design features a GAP9 MCU, an OV5647 monocular camera, a VL53L5CX Time-of-Flight (ToF) depth sensor\footnote{VL53L5CX: \url{https://www.st.com/en/imaging-and-photonics-solutions/vl53l5cx.html}}, 32~MB of onboard PSRAM\footnote{AP Memory Pseudo Static RAM: \url{https://www.mouser.at/datasheet/2/1127/APM_PSRAM_OPI_Xccela_APS256XXN_OBRx_v1_0_PKG-1954780.pdf}}, and 64~MB of Flash memory. 
In addition to the standalone operational mode, the  GAP9Shield can expand the sensor front-end of a nano-drone IoT device, i.e., a Crazyflie 2.1 system\footnote{Crazyflie 2.1: https://www.bitcraze.io/products/old-products/crazyflie-2-1}, as illustrated in Fig.~\ref{fig:iotnode}. 
Thanks to the miniaturized form factor (a diameter of \SI{10}{\centi\meter} and a total weight lower than \SI{50}{\gram}), this commercial nano-quadcopter represents a valuable IoT platform for data acquisition in small and cluttered environments, e.g., a laboratory. 

The camera of the multi-modal sensor front-end produces RGB images of QVGA resolution ($320\times240$~\si{\pixel}) at a frame rate of up to \SI{120}{\null} frames per second.
The horizontal Field-of-View (FOV) measures $57^\circ$ and the average power consumption is \SI{80}{\milli\watt} in active mode. 
The VL53L5CX ToF is a multizone depth sensor that can record $8\times8$ depth maps in a range of [\SI{2}{\centi\meter}, \SI{4}{\meter}]. 
The sensor consumes  \SI{313}{\milli\watt} on average with a throughput of up to \SI{15}{\hertz}.
The horizontal FOV is $65^\circ$. 

On the processing side, GAP9 includes a 32-bit RISC-V CPU, \SI{1.5}{\mega\byte} of on-chip RAM memory, and a wide set of peripherals to connect to the sensors or external storage memories, including a dedicated I/O DMA engine for off-chip data management. 
The single-core architecture is enriched by a software-programmable accelerator, denoted as Cluster, that is composed by 9 RISC-V CPUs and \SI{128}{\kilo\byte} of scratchpad memory. 
A second DMA engine, reserved for the Cluster, can transfer data between the scratchpad memory and the larger on-chip RAM in the background of the CPU operation.  
The cluster cores share 4 Floating-Point Units, with support for 16-bit floating-point (\texttt{BFloat16} format) calculation. 
The presence of these HW blocks extends the instruction set of the CPU with Single-Instruction-Multiple-Data (SIMD) instructions that can process $2\times$\texttt{BFloat16} vectors. 
When running at maximum clock speed of \SI{370}{\mega\hertz}, the average power consumption is typically below \SI{100}{\milli\watt}.

\begin{figure}[t]
\centering
\includegraphics[width=1\linewidth]{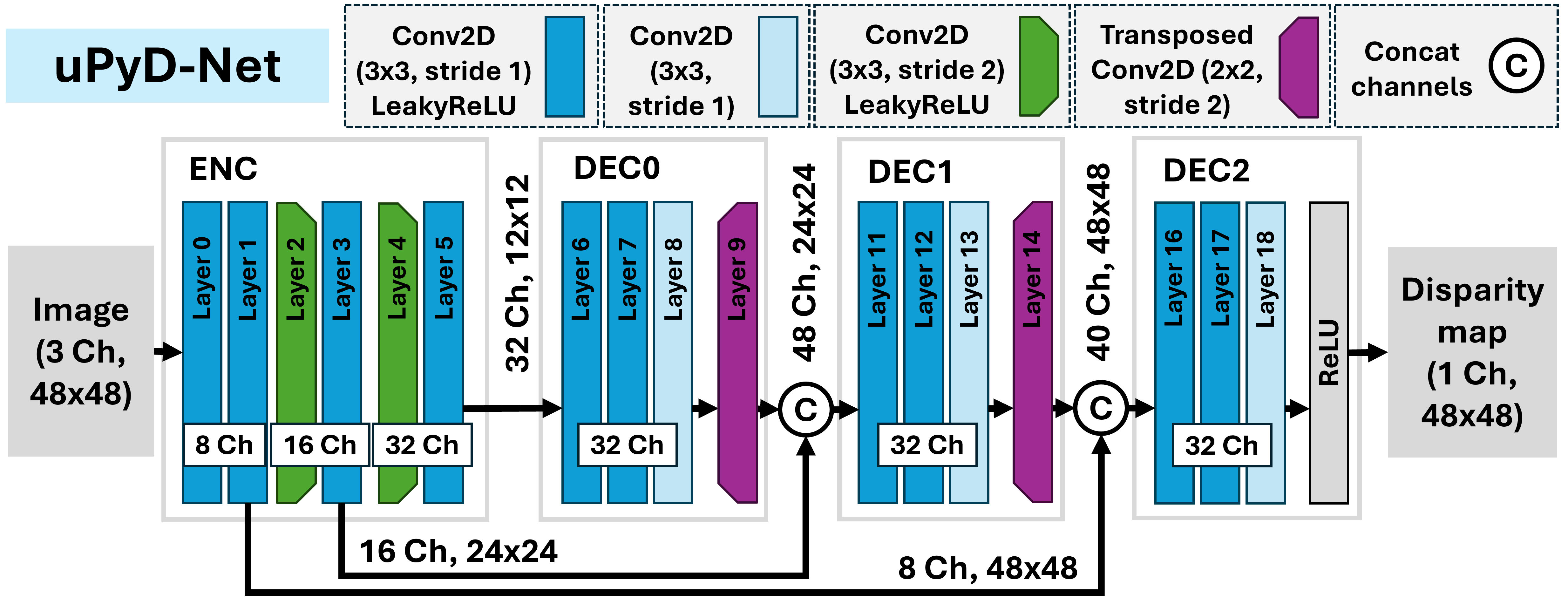}
\caption{$\mu$PyD-Net~\cite{peluso2021monocular} DNN architecture for monocular depth estimation.}
\label{fig:upydnet-archi}
\end{figure}

\begin{figure*}[t]
\centering
\includegraphics[width=1\linewidth]{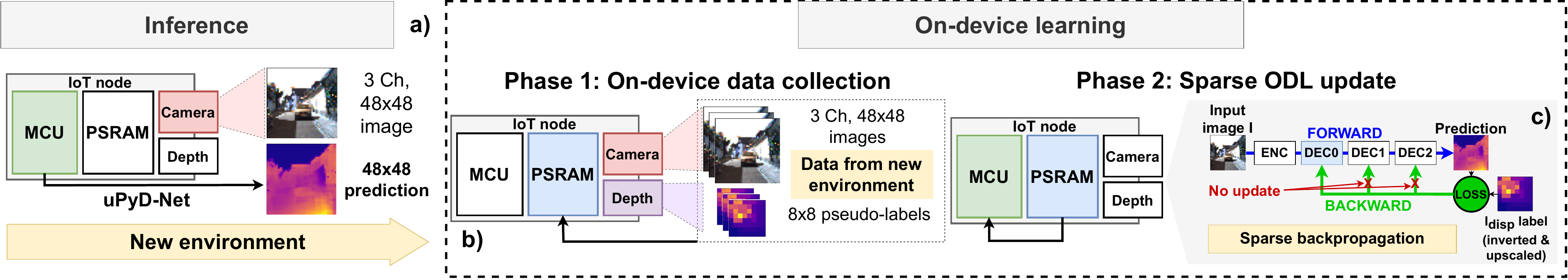}
\caption{
Diagram of the main phases of our ODL method for MDE. 
(a) The IoT node continuously runs the inference-only task onboard and predicts the depth maps from single-camera images.
In ODL mode, the system collects new data from the camera and the depth sensors (b) and then performs training on-device (c). 
In every phase, system components colored in white are not active.} 
\label{fig:iotsetup}
\end{figure*}

\section{Method}\label{sec:method}
 
In this Section, we discuss our novel On-Device Learning (ODL) system-level method aimed at improving the performance of tiny DNNs for monocular depth estimation after deployment on an IoT sensor node. 
The sensor system described in Sec.~\ref{sec:hardware} can operate in two distinct modes. 
Normally, it runs the MDE task based only on image data from the low-power OV5647 camera, while the low-resolution depth sensor is kept off.
When placed in a new environment, the system starts the ODL routine, which comprises two phases:
\begin{enumerate}
    \item{\textbf{Onboard data collection}:} 
    the system records a new set of images from the target environment and assigns pseudo-labels based on the depth information retrieved by the other sensor. 
    The new dataset is stored on the on-board PSRAM memory. 
    \item {\textbf{Sparse ODL}:}
    the DNN-based MDE model is fine-tuned on the MCU with respect to the collected data. 
    Our lightweight ODL strategy adopts a sparse update rule where only a subset of the parameters are updated during training. 
\end{enumerate}

In the following, we detail the two main phases of our method after describing our baseline sensor system for MDE. 
Fig.~\ref{fig:iotsetup} illustrates these components.

\subsection{Baseline MDE System} \label{sec:baseline}

Our work builds on the state-of-the-art $\mu$PyD-Net~\cite{peluso2021monocular} for on-device MDE. 
$\mu$PyD-Net, which is illustrated in Fig.~\ref{fig:upydnet-archi}, is a U-Net-like Convolutional Neural Network (CNN) that generates depth information with the same resolution as the input visual image.
The model includes an encoder (ENC) and three decoders, which are indicated as DEC0, DEC1, and DEC2.  
The ENC block features six 2D convolution layers followed by LeakyReLU activations. 
Every encoder layer expands the number of channels of its input tensor data and progressively halves the spatial resolution. 
In contrast, every decoder block upscales by 2$\times$ the spatial dimension of the inputs using transposed convolutions (TrConv2D).
The output features of the DEC0 and DEC1 blocks are concatenated with the encoder's intermediate results, which are propagated via the residual connections, before feeding the following decoder stage.

In the original work, $\mu$PyD-Net is trained to predict a $48 \times 48$~\si{\pixel} disparity map $\mathcal{I}_{disp}$, which reflects the pixel-wise displacement between the left and right images of the training data (KITTI). 
A depth map, i.e., a $48 \times 48$~\si{\pixel} image that encodes the pixel-level distance (in \SI{}{\meter}) of the scene from the camera, is then obtained as:

\begin{equation}
\label{eq:disparity}
\mathcal{I}_{dpth} = (f \cdot B) / \mathcal{I}_{disp}, 
\end{equation}
%
where $f$ and $B$ are scalar parameters indicating, respectively, the focal length of the RGB camera used to collect images and the stereo baseline distance between the centers of projection of the stereo camera pair. 
Given the use of stereo images for training, the authors of $\mu$PyD-Net employ a training loss that combines a supervised \textit{reverse Huber} (berHu)~\cite{zwald2012berhu} loss with an unsupervised \textit{appearance matching}~\cite{godard2017unsupervised} stereo re-projection loss.
The supervised loss computes the error between $\mu$PyD-Net's prediction and a disparity map derived from the left and right training images.
The unsupervised component, instead, returns a similarity score between one training image (e.g., the left) and the other image of the stereo pair after applying a warping transformation according to the predicted disparity map. 
 
Since our MDE model is intended for on-device training with monocular images, we initially train $\mu$PyD-Net on a dataset featuring single-camera images annotated with depth ground truth.  
Differently from the original work, we use only the supervised \textit{berHu}  loss, which computes the error between $\mu$PyD-Net's disparity predictions and the respective depth labels, inverted to a disparity map using Eq.\ref{eq:disparity}. 
We experimentally verified that our training recipe leads to the same accuracy as the loss function used in the original work.

After training, $\mu$PyD-Net~ is deployed on the GAP9 MCU of our system (Fig.~\ref{fig:iotsetup}-a).
In this work, we rely on the latency-optimized kernels of PULP-TrainLib~\cite{nadalini2023reduced} to estimate the latency and energy consumption of $\mu$PyD-Net. 
We feed the model with single images featuring CHW data layout.
The DNN processing is performed layer-by-layer using the Cluster cores. 
%
%
The weights and intermediate tensors data, which are initially stored in the on-chip \SI{1.5}{\mega\byte} memory, are copied to the Cluster scratchpad memory before invoking the layer-wise functions. 
The results of the computation are then stored back to the on-chip RAM memory for later use using the DMA. 

Thanks to the optimized deployment library, our IoT node can predict at runtime the $\mathcal{I}_{dpth}$ output by feeding the network with RGB images captured by the Omnivision OV5647 monocular camera. 
To feed $\mu$PyD-Net, every $320 \times 240$~\si{\pixel} image is downsampled to $48 \times 48$ resolution using nearest neighbor interpolation. 
While working in this pure inference mode, the low-resolution depth sensor (the VL53L5CX ToF) is turned off to reduce the energy consumption of the node (the depth sensor consumes $\sim2 \times$ more than the combination of camera and MCU).

\subsection{On-device Data Collection} \label{sec:collect}
When placed in a new environment, 
the system records a new image dataset for on-device learning using the available sensors. 
More in detail, every image recorded by the camera is associated with the low-resolution depth map output by the VL53L5CX ToF sensor.
Note that, however, this \textit{pseudo-label} 
%
%
is not equivalent to the ground truth depth because of two factors. 
First, the pseudo-labels are partial, as the VL53L5CX sensor is limited to a maximum range of 4~m, often constrained to $\sim$2~m in real-world conditions due to non-ideal lighting and surface reflectance. 
During fine-tuning, we address this range issue by avoiding the loss computation for every pixel marked as invalid by the depth sensor. 

Second, the camera features a narrower FOV than the depth sensor, which causes a mismatch between the images' content and the respective measured depth. 
Instead of aligning the two sensors by warping the depth map to match the camera's narrower FOV—which would discard useful depth information corresponding to edge regions of the camera image—we retain the full depth map to preserve all potentially relevant spatial data.
This approach produces noisy labels, which require the model to learn the sensor alignment from the data itself, during on-device fine-tuning. 

The new set of labeled images is stored in the on-board PSRAM memory (Fig.~\ref{fig:iotsetup}-b).
Thanks to the total capacity of 32~MB, our system can retain up to 4587 $48\times48$ RGB images (\SI{6.9}{\kilo\byte} each after downsampling) and the associated $8 \times 8$ depth maps (\SI{64}{\byte} each). 
Given the throughput of the slowest sensor (\SI{15}{\hertz} for the VL53L5CX vs. \SI{120}{\hertz} for the camera), the maximum amount of data is collected in \SI{305.8}{\second}.

\subsection{Sparse On-device Learning Update}\label{sec:sparse_update}

\subsubsection{Backpropagation-based learning for MDE}

We use the backpropagation algorithm to fine-tune $\mu$PyD-Net on-device.
The backpropagation algorithm operates in two steps, namely the \textit{forward}, computing the DNN disparity prediction with respect to an input image $\mathcal{I}$, and the \textit{backward}, which calculates the update values for the weights to be fine-tuned. 
During the forward step, the intermediate activations of the layers to be updated (e.g., convolutions) and of the activation functions across which the error is propagated (e.g., ReLU) are retained in memory for the backward pass. 

After the forward step, we use the same loss function of the initial training to calculate the error between the model prediction ($48 \times 48$~\si{\pixel}) and the noisy label, collected in field ($8 \times 8$~\si{\pixel}).
To compute the loss, the label needs to be transformed. 
First, we invert the label to a disparity map, using Eq.~\ref{eq:disparity}.
Then, we upscale it using a bilinear interpolation to $48 \times 48$~\si{\pixel}. 
The backward step then backpropagates the error signal up to the first layers with trainable parameters.

The obtained gradient values are eventually averaged over a mini-batch of samples (image and depth), and then accumulated into the weight values after applying the learning rate. 
Since $\mu$PyD-Net does not include any batch-dependent layer, e.g., BatchNorm, the backpropagation algorithm can operate sample-by-sample, in a streaming fashion.


\subsubsection{On-Device Learning}
The backpropagation-based training task runs entirely on the GAP9 MCU, without involving any off-device (server) computing resource (Fig.~\ref{fig:iotsetup}-c). 
To improve computation and memory efficiency, we select the \texttt{BFloat16} floating-point datatype to represent weights, gradients, and intermediate feature values, similarly to the inference phase. 
This data format allows the use of SIMD-accelerated software kernels from PULP-TrainLib, which can lead to up to 1.9$\times$ speed-up vs. the 32-bit floating point counterpart~\cite{nadalini2023reduced} without sacrificing accuracy~\cite{kalamkar2019study}.

\begin{figure}[t]
\centering
\includegraphics[width=1\linewidth]{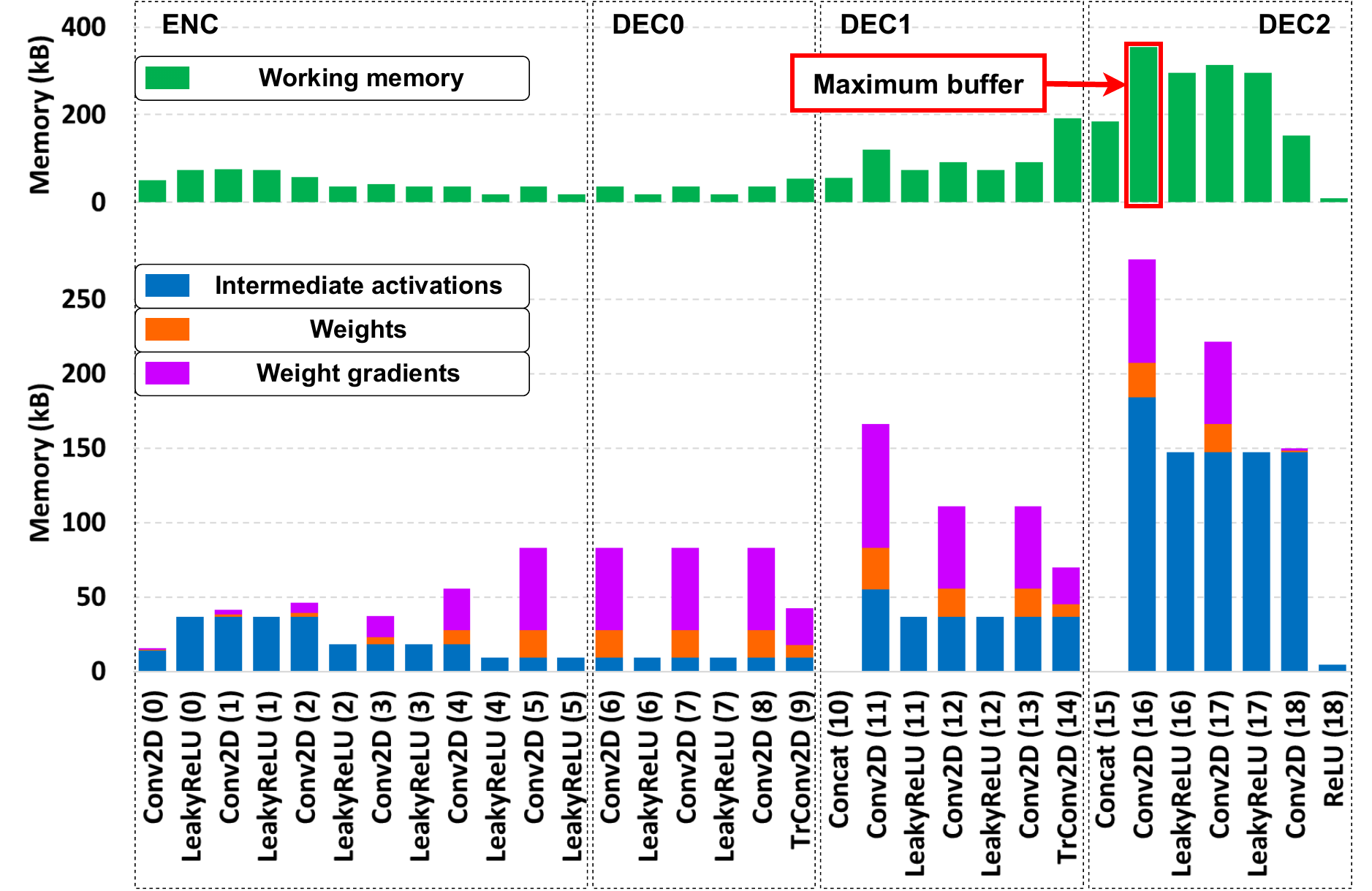}
\caption{Layer-wise breakdown of $\mu$PyD-Net memory cost, contributing to the total fine-tuning memory.}
\label{fig:upydnet}
\end{figure}

Our implementation runs the $\mu$PyD-Net forward and backward passes layer-by-layer using the cluster accelerator of GAP9 and uses the \SI{1.5}{\mega\byte} on-chip memory as the main computation memory.
Fig.~\ref{fig:upydnet} illustrates the total memory requirements to train $\mu$PyD-Net on GAP9. 
The total fine-tuning memory (2.56~MB) is composed by two components, i.e., a working memory buffer to compute every layer's activations and gradients, and a storage buffer for the backward step data. 

More specifically, the working memory buffer is dimensioned for computing the largest among all layers' activations and gradients, and is shared by the forward and backward steps of all layers.
The storage buffer, instead, contains the model weights, the intermediate activations, and the weight gradients for the layers to be updated.
This last cost also accounts for the additional memory required by the optimization process, e.g., the buffers for the estimation of the statistical moments in the case of the Adam optimizer \cite{kingma2014adam}. 
%
During ODL, while the working memory buffer is always kept on-chip, storage buffer data that exceeds the \SI{1.5}{\mega\byte} on-chip memory is stored in the on-board PSRAM. 
When requested, such data is accessed using the I/O DMA, in parallel with CPU operations. 

In the U-Net-like~\cite{azad2024medical} architecture of $\mu$PyD-Net, introduced in Sec.~\ref{sec:baseline}, DEC2 impacts the most on the fine-tuning memory cost.
%
The size of the working memory buffer (354.9~KB) is determined by the first convolution of DEC2, which features the largest input activation of the model (40 channels with $48 \times 48$ spatial size).
Similarly, the size of the storage memory buffer (2.2~MB) is determined for the 43.1\% by DEC2 (947.5~kB), while ENC, DEC0, and DEC1 contribute to its 18.6\%, 14.1\% and 24.2\%, respectively.
Of the whole storage buffer, 1.3~MB (60.8\%) are required to store the intermediate activations of the model, while the remaining 39.2\% stores the weights and their related gradients. 
Such requirements are unevenly distributed among the blocks: while DEC2 contains 58.3\% of all activations (778.8~kB), it possesses only 19.6\% of the weights of the model.
The other blocks (ENC, DEC0, DEC1, respectively), instead, require 19.7\%, 17.9\%, and 4.1\% of the stored activation cost.
At the same time, the central blocks (DEC0, DEC1) contain the majority of the weights (29.6\% and 33.9\%), while the ENC possesses only its 16.9\%.

\subsubsection{Sparse Update}

We use a sparse update approach to fine-tune only a subset of the $\mu$PyD-Net parameters. 
Our proposed layer selection logic aims to minimize the memory requirements vs. the full model update while preserving final accuracy in a regression task, i.e., MDE. 
With respect to typical DNNs used for classification tasks, which show the lowest cost when retraining only the last layers \cite{paissan2024structured}, we make the fundamental observation that our U-Net-like $\mu$PyD-Net presents the highest memory cost in correspondence to the last layers. 
Furthermore, the computation of DEC2's weight gradients alone constitutes 28.2\% of the total cost for a backward step, making the weight update for the last layers the most computationally demanding, while early blocks of layers (ENC, DEC0) only represent 3.4\% and 3.7\% of such cost.
%

Therefore, our memory-driven sparse fine-tuning strategy updates only DEC0, which constitutes the memory-minimal solution due to the small footprint of its activations. 
In this configuration, the total fine-tuning memory is reduced to 1.2~MB ($2.2\times$ less than a full update). 
Specifically, the storage buffer (835~kB) accounts for 310.2~kB corresponding to DEC0 activations and gradients, 373.2~kB for the input activations of the ReLU and LeakyReLU layers in DEC1 and DEC2 (required for error propagation), and 151.5~kB for the weights of the remaining network blocks.
This setup reduces by 24\% the memory cost, compared to the standard fine-tuning approach, which updates only the final layers (DEC2 for $\mu$PyD-Net). 
In this conventional scheme, the storage buffer alone requires 1.1~MB, comprising 947.5~kB for the activations, weights, and gradients of DEC2, and an additional 173~kB for the weights of the other network blocks used during the forward pass.


In Sec.~\ref{sec:sparse_update_analysis}, we experimentally show that our memory-efficient sparse configuration also retains the same accuracy score compared to retraining the full model. 
This result is in line with the observation of previous studies that observed no accuracy drops when incrementally updating only intermediate layers of DNNs tailored for classification tasks \cite{kwon2024tinytrain} and U-Net models for ultrasound image segmentation~\cite{amiri2020fine}.
%
In compliance with the limited amount of RAM of our IoT platform, we also demonstrate that our sparse update scheme is robust to a significant reduction in the number of samples available for fine-tuning on-device.

\section{Experimental Data Protocol}
\label{sec:training}


We define a new data protocol to validate our ODL method for MDE. 
Using data from available public datasets, we initially mimic an ideal sensor setup with a perfect matching between the FOVs of the camera and the depth sensor. 
More in detail, the baseline $\mu$PyD-Net is trained on  TartanAir~\cite{wang2020tartanair}, and then we fine-tune it on data from the KITTI~\cite{Geiger2013IJRR} or NYUv2~\cite{silberman2012indoor} datasets.  
\textbf{TartanAir} is a synthetic dataset comprising 4 TB of photorealistic data from structured urban, indoor, and unstructured natural environments, annotated with dense depth maps and featuring a resolution of $640 \times 480$~\si{\pixel}. 
\textbf{KITTI} is a popular computer vision benchmark featuring stereo images with an average resolution of $1242 \times 375$~\si{\pixel}, coupled with point-cloud LIDAR depth measurements, capturing scenes from urban, suburban, and rural environments. 
\textbf{NYUv2} is a popular Microsoft Kinect benchmark comprising more than 400,000 RGB $640 \times 480$ images, annotated with dense depth maps, collected as video sequences from various indoor environments, such as kitchens, living rooms, bedrooms, and bathrooms.

TartanAir, KITTI, and NYUv2 are pre-processed by cropping centered portions of every image and label, which are then downsampled to $48 \times 48$ resolution with nearest neighbor interpolation to feed $\mu$PyD-Net. 
To keep the same downsampling ratio for all datasets, we apply the cropping size adopted for KITTI ($360 \times 360$~\si{\pixel}).
This size is chosen to always crop square portions of images and labels in such dataset, since several samples feature a vertical resolution less than the average one (375~\si{\pixel}).

In our protocol, we pre-train $\mu$PyD-Net on TartanAir, extracting 64358 training images from 3 environments (\textit{AbandonedFactory, Neighborhood, SeasonsForest}), using $48 \times 48$ labels.
%
We evaluate our fine-tuning on KITTI and NYUv2. 
For the KITTI dataset, we adopt the Eigen split~\cite{eigen2014depth}, which comprises 22600 images for training, 888 for validation, and 697 for testing.
Since KITTI features ground truth data collected with a LIDAR sensor that produces sparse measurements, we follow~\cite{peluso2021monocular} to generate dense proxy labels from stereo images, for training. 
For the NYUv2 dataset, we sample 22600 images for training, 2726 for validation, and 2948 for testing from 180 sequences of kitchens, living rooms, bedrooms, and bathrooms, each divided into 80\% data for training, 10\% for validation, and 10\% for testing.
To eventually match the specifications of the VL53L5CX depth sensor, we simulate the $8 \times 8$ depth labels used for fine-tuning by downsampling every $48 \times 48$ label of KITTI and NYUv2 with a $6 \times 6$ min-pooling filter. 


In a second phase, we assess the performance of our ODL solution on a set of real-world data, \textbf{IDSIA-$\mu$MDE}
\footnote{IDSIA-$\mu$MDE: \url{https://github.com/idsia-robotics/ultralow-power-monocular-depth-ondevice-learning}}, 
collected with our IoT setup (Sec.~\ref{sec:hardware}).
IDSIA-$\mu$MDE features 3035 labeled images for training, 1149 for validation, and 3171 for testing, collected from different perspectives of the same laboratory environment, featuring artificial lighting and a set of obstacles, such as recycle bins, robots, and gates.
Due to the non-idealities of our IoT setup (i.e., sensor FOV mismatch and limited depth sensor range), IDSIA-$\mu$MDE images are annotated with noisy labels, which we employ both for fine-tuning and testing.

\section{Results}\label{sec:results}

\subsection{Experimental Setup}\label{sec:exp_setup}


We evaluate our ODL solution using the PyTorch framework\footnote{PyTorch: https://pytorch.org/}, and we report every result as the average over 5 experiments.  
Additionally, we profile the training task running on the GAP9 MCU by measuring the latency and power consumption of the IoT system.

\textbf{Metrics.} We evaluate the accuracy of MDE prediction by employing three metrics.
We define as $\mathcal{I}_{GT}$ the depth ground truth and with $N$ the number of elements in $\mu$PyD-Net's prediction $\mathcal{I}_{dpth}$. 
The quality of pixel-wise predictions is measured using the accuracy under threshold $1.25^{K}$ ($\delta_K$). 
$\delta_K$ expresses the ratio of pixels in which the predicted depth is within a multiplicative threshold $1.25^{K}$ of the ground truth:

\begin{equation}
    \delta_K = \frac{1}{N} \sum_{i \in N} I\left\{ \max\left( \frac{\mathcal{I}_{dpth,i}}{\mathcal{I}_{GT,i}}, \frac{\mathcal{I}_{GT,i}}{\mathcal{I}_{dpth,i}} \right) < 1.25^K \right\},
\end{equation}

where $I\{\cdot\}$ is an indicator function that equals 1 if its argument is true and 0 otherwise. 
The root mean squared error ($RMSE$) measures the average prediction error, in meters, between the prediction and ground truth, with the formula:

\begin{equation}
    RMSE  = \sqrt{\frac{1}{N} \sum_{i \in N} ||\mathcal{I}_{dpth,i} - \mathcal{I}_{GT,i}||^2}
\end{equation}

The Scale Invariant Mean Squared Error in Log Space ($SiLog$) metric expresses, on a logarithmic scale, the relative scale error between the ground truth and the depth prediction. We express $SiLog$ following the formula:

\begin{equation}
    SiLog = \frac{1}{N} \sum_{i \in N} d_i^2 - \frac{1}{N^2} \left( \sum_{i \in N} d_i \right)^2
\end{equation}

Where $d_i = \log(\mathcal{I}_{dpth,i}) - \log(\mathcal{I}_{GT,i})$ and represents the pixel-wise error in log scale.
Readers may refer to~\cite{eigen2014depth} for more details on the common metrics used to evaluate MDE.
For the public datasets, we test the accuracy of $\mu$PyD-Net by upsampling with nearest neighbor interpolation its $48 \times 48$ predictions to $360 \times 360$, inverted to a depth map using Eq.~\ref{eq:disparity}, and by comparing them with $360 \times 360$ portions of the ground truth. 
For the real-world data, the accuracy of $\mu$PyD-Net is tested by comparing its $48 \times 48$ predictions, inverted with Eq.~\ref{eq:disparity}, with the $8 \times 8$ noisy labels, upscaled to $48 \times 48$ resolution with nearest-neighbor interpolation.

\textbf{Hyperparameters Settings.} Following~\cite{poggi2018towards}, we set the optimizer to Adam ($\beta_1 = 0.9, \beta_2 = 0.999, \epsilon = 10^{-8}$), the batch size to 16, and the loss function to a \textit{reverse Huber} (berHu)~\cite{zwald2012berhu}, both for the initial training and fine-tuning phases. 
The learning rate is set to $10^{-4}$ for models trained from scratch and $10^{-3}$ for fine-tuned models.
We set the maximum epochs for pre-training (TartanAir) to 200, and to 120 for fine-tuning on both public datasets (KITTI and NYUv2) and data collected in the field.
An early stopping mechanism selects the model belonging to the epoch where the validation loss is minimized. 
We augment the training data with three photometric transformations, i.e., random gamma correction [0.8, 1.2], random brightness variation [0.5, 2.0], channel-wise color change [0.8, 1.2], and horizontal flipping.

\subsection{Full Update Analysis}\label{sec:full_update_analysis}

In this section, we analyze the accuracy of our $\mu$PyD-Net when fine-tuning all the parameters of the model, i.e., the full update scheme.
As additional terms of comparison, in Table~\ref{tab:full_update}, we also provide the results of the baseline (non-fine-tuned) and two dummy predictors.
Baseline and fine-tuned models are pre-trained on the TartanAir dataset (i.e., from a simulator); instead, the dummy predictors predict the pixel-wise average depth of all testing samples.
For each row of Table~\ref{tab:full_update}, we report the testing dataset on either KITTI or NYUv2, also used for fine-tuning.
For both fine-tuned models and dummy predictors, we evaluate two label configurations: an ideal 48$\times$48 and 8$\times$8 px labels, where the latter represents the actual labels' resolution on our IoT node. 
Finally, in our analysis, we also consider both fine-tuning with all the dataset samples, i.e., \SI{22.6}{\kilo\nothing}, and with a subset constrained by the available device's PSRAM (\SI{32}{\mega\byte}), resulting in \SI{2.3}{\kilo\nothing} and \SI{4.5}{\kilo\nothing} samples, for the 48$\times$48 and 8$\times$8 px labels, respectively.

\begin{table}[t]
\centering
\caption{Accuracy evaluation of our fine-tuned models (full update) vs. no fine-tuned baseline and dummy predictors. All models are pre-trained on the TartanAir dataset.}
\def\arraystretch{1.3}
\resizebox{\linewidth}{!}{
\begin{tabular}{lccccc}
\hline
\multirow{2}{*}{\textbf{Model}} & \multirow{2}{*}{\textbf{Dataset}} & \textbf{Fine-tuned} & \textbf{\pmb{$\delta_1$}} & \textbf{RMSE} & \textbf{SiLog} \\
& & \textbf{(Samples [\SI{}{\kilo\nothing}])} & \textbf{[\%] \pmb{$\uparrow$}} & \textbf{[\SI{}{\meter}] \pmb{$\downarrow$}} & \textbf{\pmb{$\downarrow$}} \\ 
\hline
Baseline & KITTI & \xmark~(--) & 17.8 & 19.5& 6.8 \\
\textit{(no fine-tuning)} & NYUv2 & \xmark~(--) & 3.6 & 74.9 & 8.7 \\ 
\hline
\multirow{4}{*}{\begin{tabular}[c]{@{}l@{}}Full update \\ \textit{($48 \times 48$ labels)}\end{tabular}} & \multirow{2}{*}{KITTI} & \cmark~(22.6) & 81.9 & 7.2 & 2.0 \\
& & \cmark~(2.3) & 80.8 & 7.4 & 2.1 \\ 
\cline{2-6} 
& \multirow{2}{*}{NYUv2} & \cmark~(22.6) & 51.7 & 0.5 & 2.5 \\
& & \cmark~(2.3) & 49.2 & 0.5 & 2.6 \\ 
\hline
\multirow{4}{*}{\begin{tabular}[c]{@{}l@{}}Full update \\ \textit{($8\times8$ labels)}\end{tabular}} & \multirow{2}{*}{KITTI} & \cmark~(22.6) & 63.2 & 9.3 & 2.3 \\
& & \cmark~(4.5) & 62.3 & 9.2 & 2.3 \\ 
\cline{2-6} 
& \multirow{2}{*}{NYUv2} & \cmark~(22.6) & 50.1 & 0.5 & 2.5 \\
& & \cmark~(4.5) & 49.3 & 0.5 & 2.6 \\ 
\hline
\multirow{2}{*}{\begin{tabular}[c]{@{}l@{}}Dummy predictor \\ \textit{($48 \times 48$ labels)}\end{tabular}}& KITTI & \xmark~(--) & 75.3 & 9.8 & 2.8 \\
& NYUv2 & \xmark~(--) & 40.2 & 0.6 & 2.6 \\ 
\hline
\multirow{2}{*}{\begin{tabular}[c]{@{}l@{}}Dummy predictor \\ \textit{($8\times8$ labels)}\end{tabular}}& KITTI & \xmark~(--) & 45.7 & 14.3 & 3.3 \\
& NYUv2 & \xmark~(--) & 39.9 & 0.6 & 2.6 \\
\hline
\end{tabular}
\label{tab:full_update}
}
\end{table}

\begin{figure}[t]
\centering
\includegraphics[width=1\linewidth]{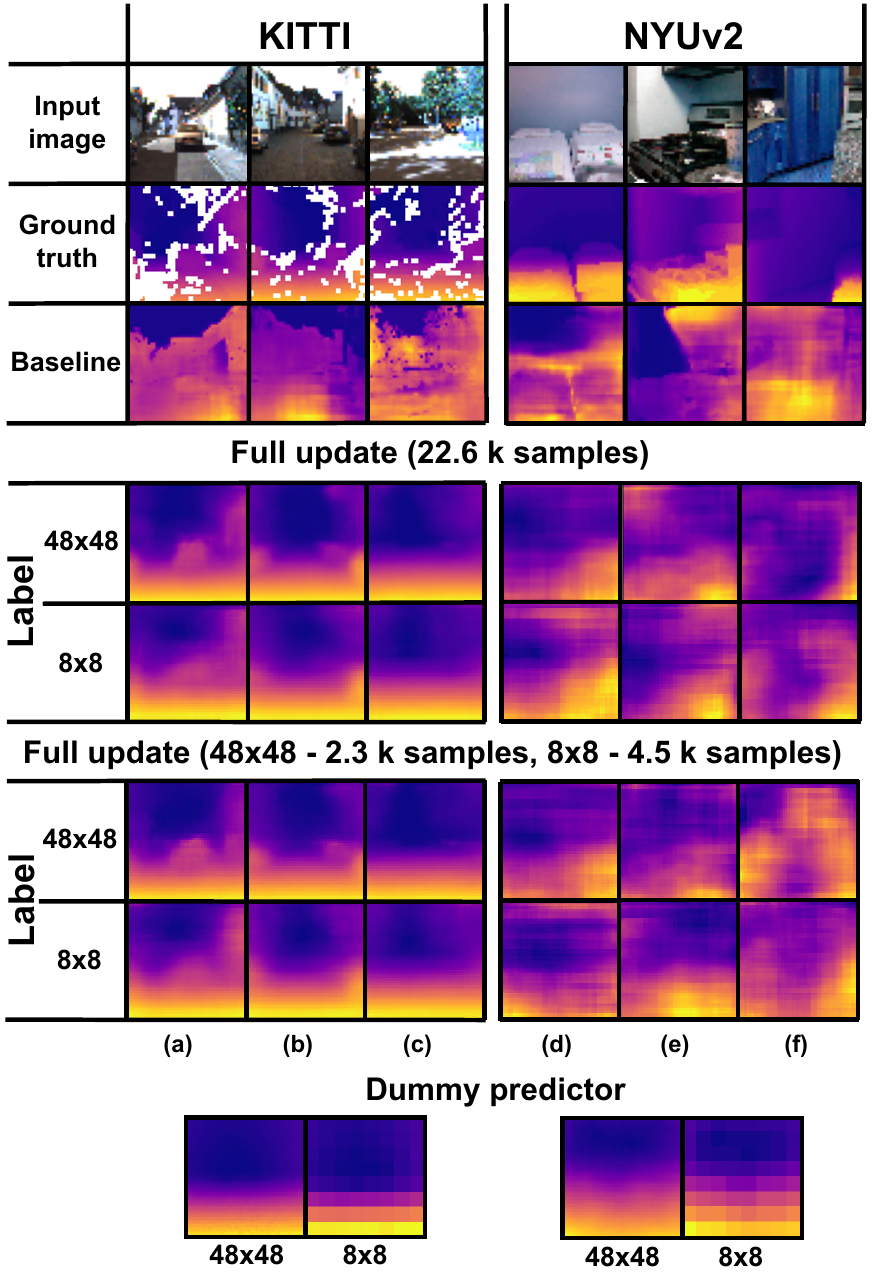}
\caption{Images, ground truths and $\mu$PyD-Net predictions.}
\label{fig:depthpredictioncomp}
\end{figure}

As can be observed from Table~\ref{tab:full_update}, the baseline model struggles to generalize, with a limited 17.8\% and 3.6\% $\delta_1$, and a RMSE of \SI{19.5}{\meter} and \SI{74.9}{\meter}, for the KITTI and NYUv2 datasets, respectively.
This performance highlights the domain shift problem between training in simulated environments and real-world testing ones.
The best $\delta_1$ accuracy comes with the models fine-tuned with 48$\times$48 px labels on the full dataset (i.e., \SI{22.6}{\kilo\nothing} samples).
In this case, $\mu$PyD-Net achieves +64.1\% and +48.1\% $\delta_1$ on KITTI and NYUv2, respectively, compared to the baseline.
Instead, with 8$\times$8 px labels and the full fine-tuning datasets, the relative improvement w.r.t. the baseline reduces to +45.4\% and +46.5\% $\delta_1$, on KITTI and NYUv2, respectively.

While for NYUv2, we observe a similar fine-tuning accuracy with both label resolutions, $\mu$PyD-Net achieves 18\% lower accuracy in the case of KITTI, using $8 \times 8$ labels.
This effect derives from downsampling the labels to emulate the VL53L5CX sensor in outdoor environments.
In this case, regions featuring large depth variance in neighboring pixels are downsampled to the nearest depth, distorting the depth distribution (e.g., a region containing both a car and a portion of the sky is collapsed to the car's depth).
Moving to the smaller fine-tuning datasets, the models in both label configurations show a negligible loss in all metrics, i.e., a reduction of only 2.5\% at most in $\delta_1$, for NYUv2 with 48$\times$48 px labels.
Finally, considering the dummy predictors introduced as sanity checks, they consistently score better than the baseline but significantly worse than all fine-tuned models in all configurations.
These results demonstrate the full update's effectiveness in limiting the domain shift problem even when employing small labels and a limited number of samples.

In Fig.~\ref{fig:depthpredictioncomp}, we report a qualitative analysis of the various models presented in Table~\ref{tab:full_update} with three samples for each testing dataset.
As previously introduced in Sec.~\ref{sec:training}, ground truths for the KITTI dataset (Fig.~\ref{fig:depthpredictioncomp}-a/b/c) are produced with a stereo camera, which suffers more from occlusions, resulting in missing ground truth depths (white pixels).
Instead, the NYUv2 dataset, employing a 640$\times$480 px Kinect depth sensor, produces dense labels (Fig.~\ref{fig:depthpredictioncomp}-d/e/f).
The baseline model is strongly biased by the pixels' brightness, predicting high-depth values for pale-color pixels.
Conversely, models fine-tuned with all \SI{22.6}{\kilo\nothing} samples, independently from their labels' resolution, accurately detect obstacles in KITTI, clearly separating cars from backgrounds.
For NYUv2, instead, they capture the scene details less accurately while correctly segmenting most areas with homogeneous depth, especially in the lower part of the images.

The models fine-tuned with \SI{2.3}{\kilo\nothing} and \SI{4.5}{\kilo\nothing} samples and 48$\times$48 px labels are less precise in predicting objects' edges, in particular for those on the forefront, while they are pretty accurate for far-distant objects and backgrounds.
Instead, the models fine-tuned with 8$\times$8 px labels produce blurrier predictions, resulting from the bilinear upsampling used to generate these small labels, which creates smooth transitions in the fine-tuning's upscaling process.
Finally, considering the two dummy predictors that output only the pixel-wise average of the testing datasets, they capture the correlation between the background in the upper part of the images (higher depth) and the foreground (lower depth) in their lower part.

\subsection{Sparse Update Analysis}\label{sec:sparse_update_analysis}

\begin{figure}[t]
\centering
\includegraphics[width=1\linewidth]{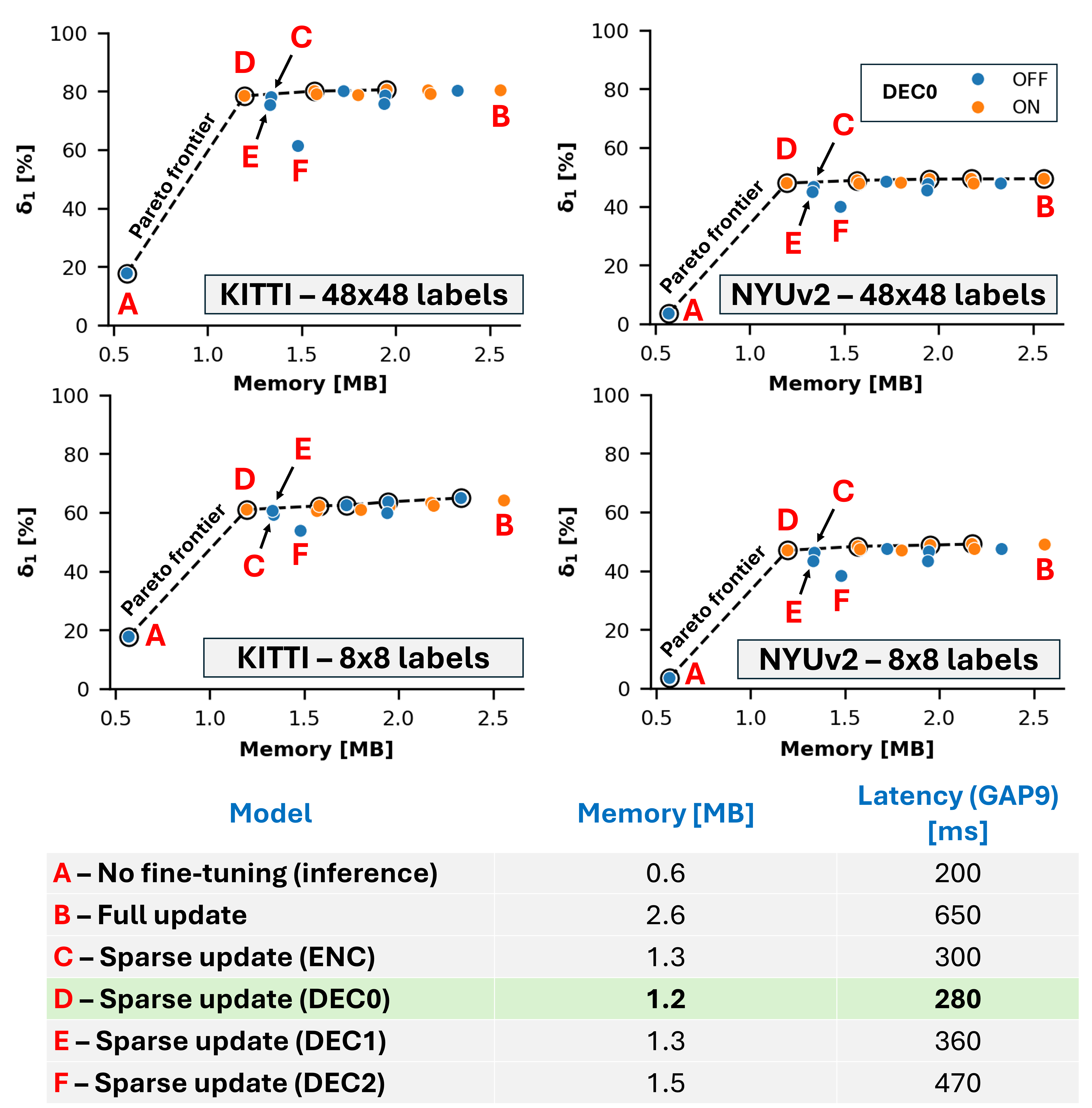}
\caption{Sparse update $\delta_1$ accuracy vs. memory for KITTI and NYUv2. We highlight the 4 most meaningful configurations.}
\label{fig:sparseupdate}
\end{figure}

\begin{table}[t]
\centering
\caption{Accuracy evaluation with the sparse update (configuration D from Fig.~\ref{fig:sparseupdate}) of our fine-tuned models.}
\def\arraystretch{1.3}
\resizebox{\linewidth}{!}{
\begin{tabular}{lccccc}
\hline
\multirow{2}{*}{\textbf{Model}} & \multirow{2}{*}{\textbf{Dataset}} & \textbf{Fine-tuned} & \textbf{\pmb{$\delta_1$}}& \textbf{RMSE} & \textbf{SiLog}\\
& & \textbf{(Samples \SI{}{[\kilo\nothing]})} & \textbf{[\%] \pmb{$\uparrow$}} & \textbf{[\SI{}{\meter}] \pmb{$\downarrow$}} & \textbf{\pmb{$\downarrow$}} \\ \hline
\multirow{4}{*}{\begin{tabular}[c]{@{}l@{}}Sparse update\\ (\textit{$48 \times 48$ labels})\end{tabular}} & \multirow{2}{*}{KITTI}& \cmark~(22.6) & 79.3 & 7.8 & 2.2 \\
& & \cmark~(2.3)& 78.5 & 7.9 & 2.3 \\ \cline{2-6} 
& \multirow{2}{*}{NYUv2}& \cmark~(22.6) & 48.7 & 0.5 & 2.6 \\
& & \cmark~(2.3)& 48.0 & 0.5 & 2.6 \\ \hline
\multirow{4}{*}{\begin{tabular}[c]{@{}l@{}}Sparse update\\ (\textit{$8\times8$ labels})\end{tabular}} & \multirow{2}{*}{KITTI}& \cmark~(22.6) & 61.0 & 9.5 & 2.3 \\
& & \cmark~(4.5)& 61.0 & 9.4 & 2.4 \\ \cline{2-6} 
& \multirow{2}{*}{NYUv2}& \cmark~(22.6) & 47.2 & 0.5 & 2.6 \\
& & \cmark~(4.5)& 47.0 & 0.5 & 2.6 \\ \hline
\end{tabular}
\label{tab:sparse_update}
}
\end{table}

This section evaluates trade-offs between fine-tuning memory cost, latency, and $\delta_1$ accuracy of various sparse update schemes.
Fig.~\ref{fig:sparseupdate} reports such results for all 16 sparse update schemes by combining all the possible configurations of the 4 blocks of $\mu$PyD-Net, i.e., a block-wise combination of ENC, DEC0, DEC1, and DEC2.
Reported latency is estimated by measuring single layers of $\mu$PyD-Net on the GAP9 MCU.
As a term of comparison, we also include the non-fine-tuned baseline models, configuration A in Fig.~\ref{fig:sparseupdate}.
Configuration B, i.e., full update, represents the scheme where all 4 blocks are used for fine-tuning their weights and biases.
Configurations from C to F refer to those where only one block of $\mu$PyD-Net is fine-tuned.
Like in Sec.~\ref{sec:full_update_analysis}, all the models are pre-trained on the TartanAir dataset, while the fine-tuning employs KITTI and NYUv2 datasets with 48$\times$48 and 8$\times$8 px labels.
In the former case, the models are fine-tuned on 2260 images and the latter on 4520 ones to respect the PSRAM capacity imposed by our IoT device (see Sec.~\ref{sec:hardware}).
Finally, Fig.~\ref{fig:sparseupdate} highlights the Pareto frontiers of the memory vs. accuracy, i.e., dashed line with black circled configurations.

All update schemes that imply fine-tuning one of the first three blocks of our $\mu$PyD-Net achieve similar $\delta_1$ accuracy, with a standard deviation of only 3.6\%, over all datasets and labels.
The only exceptions are represented by configurations A and F; no fine-tuning is applied in the former, while only the last block (DEC2) is fine-tuned in the latter.
In CNNs, shallower layers extract low-level features, such as edges and other geometric traits, while deeper ones extract more abstract features specific to the task addressed~\cite{zeiler2014visualizing}.
Therefore, in the case of domain shift due to changes in the testing dataset, like in our case, the model can benefit more from fine-tuning its geometrical interpretation (shallower layers) instead of updating deeper layers responsible for the task-specific features, as the task remains the same, i.e., depth estimation. 


$\mu$PyD-Net requires \SI{0.6}{\mega\byte} of memory to store weights and biases for the inference (i.e., forward pass).
This number grows, at least, to \SI{1.2}{\mega\byte} for the cheapest sparse update configuration D (2.1$\times$ less than the full update), making it possible to fit in the GAP9's memory. 
Configuration D is also the fastest in the sparse update backpropagation pass, requiring only \SI{80}{\milli\second} per sample.
Compared to the other update schemes on the Pareto frontier, this memory-optimal configuration D is also, at least, 14.3\% faster.
This results from the limited number of parameters updated in DEC0 (29.6\% of the total), and the possibility of stopping the backpropagation at layer 13, as there is no need to backpropagate the gradient to the ENC block.

Relative changes in the $\delta_1$ accuracy between testing datasets and labels' resolution, also in this case of sparse updates, follow similar trends as those highlighted for the full update scheme (Sec.~\ref{sec:full_update_analysis}), achieving the highest $\delta_1$ accuracy for the KITTI dataset with 48$\times$48 px labels, peaking at 80.6\% and with a lower score on the NYUv2 dataset when fine-tuned with 8$\times$8 px labels which is, at best, 49.2\%  $\delta_1$ accuracy.
For all these considerations on memory cost, latency, and $\delta_1$ accuracy, we mark configuration D as the best trade-off, showing the smallest memory footprint and fine-tuning latency, paired with an overall accuracy on par with the other sparse update schemes.
 


In Table~\ref{tab:sparse_update}, we present an in-depth analysis of configuration D of Fig.~\ref{fig:sparseupdate}.
We present the regression performance of this sparse update scheme for both KITTI and NYUv2 datasets (both for fine-tuning and testing) and both 48$\times$48 and 8$\times$8 px labels.
Additionally, like in Sec.~\ref{sec:full_update_analysis}, we also consider the availability of the entire fine-tuning dataset, i.e., \SI{22.6}{\kilo\nothing} samples or a small subset fitting the limited PSRAM memory of our IoT device.
All major relative trends of the full update analysis are also confirmed in this sparse update case.
For example, the best performing configuration is the one with 48$\times$48 px labels and the entire fine-tuning dataset, and reducing the number of fine-tuning samples marginally affects performance, e.g., the most significant drop in the $\delta_1$ accuracy is lower than 0.8\%.
Similarly, the performance drop when using the smaller labels is significant only for the KITTI dataset (i.e., 18.3\% accuracy loss) but not for the NYUv2 one, i.e., 1.5\% at most, like for the full update case and the same reasons.

\subsection{Sensitivity to Image Resolution}

\begin{figure}[t]
\centering
\includegraphics[width=1\linewidth]{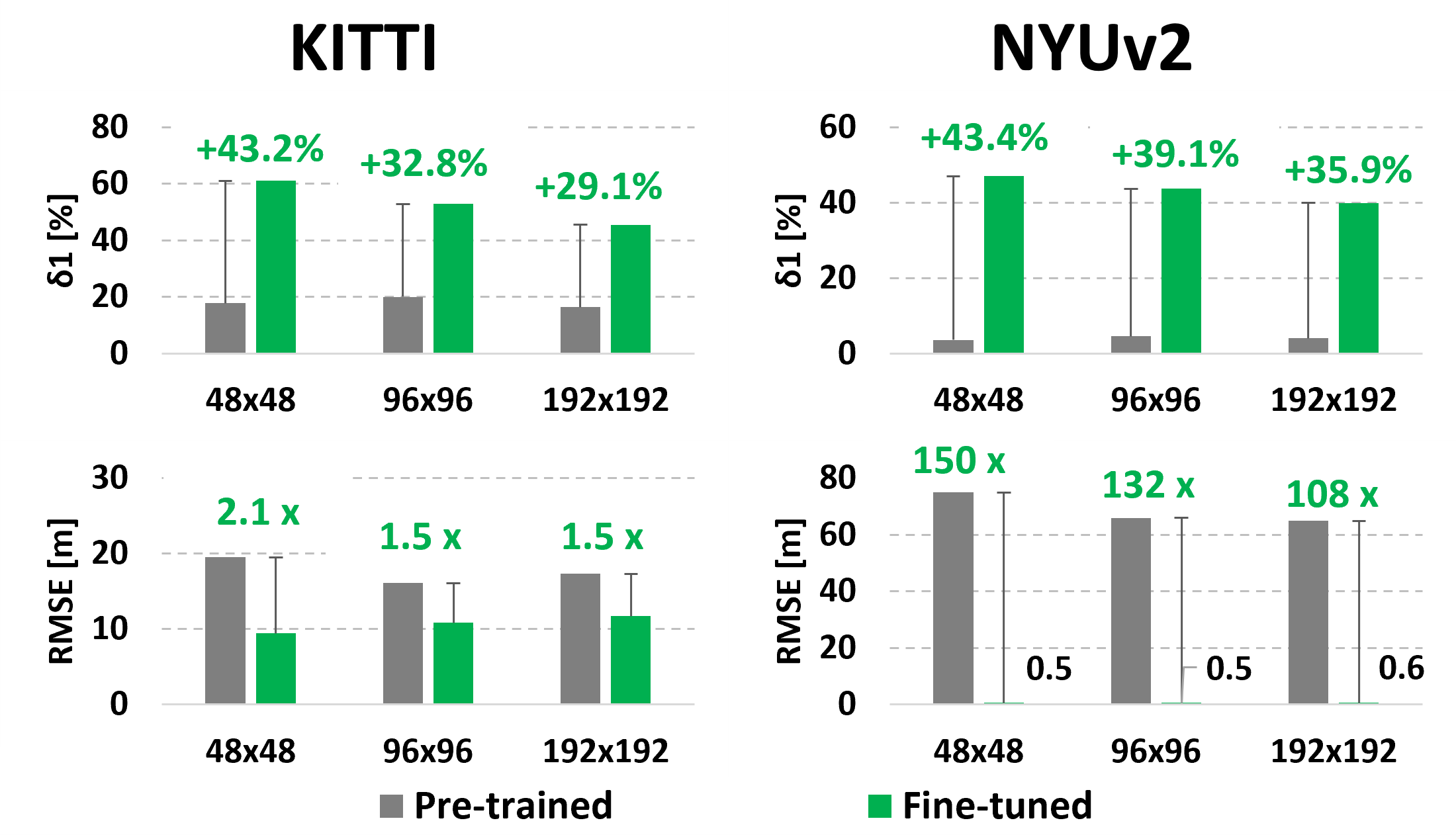}
\caption{Sensitivity of ODL to variable input resolution ($8 \times 8$ labels).}
\label{fig:multiresolution}
\end{figure}

In this section, we evaluate the sensitivity of our on-device fine-tuning strategy to input resolution, exploring its applicability to systems with higher computational resources. 
To this end, we fine-tune our pre-trained $\mu$PyD-Net on the KITTI and NYUv2 datasets, also using images with increasing resolutions of $96 \times 96$ and $192 \times 192$, corresponding to $2\times$ and $4\times$ the $48 \times 48$ resolution used in our previous experiments. 
During fine-tuning, $\mu$PyD-Net is supervised with $8 \times 8$ labels, upscaled to the target resolution via bilinear interpolation. 
Each experiment is carried out for 120 epochs, employing the most memory-efficient sparse update scheme (i.e., updating only DEC0), and uses 4.5~k training samples. The results, summarized in Fig.~\ref{fig:multiresolution}, are compared against fine-tuning with $48 \times 48$ labels.

Consistent with Table~\ref{tab:full_update}, the pre-trained model performs poorly on higher resolutions, achieving less than 20\% $\delta_1$ accuracy on KITTI and under 5\% on NYUv2. 
Conversely, when fine-tuned with upscaled $8 \times 8$ labels, the model exhibits substantial improvements across all resolutions, increasing its $\delta_1$ by +43.2\%, +32.8\%, and +29.1\% for $48 \times 48$, $96 \times 96$, and $192 \times 192$ inputs on KITTI, and by +43.4\%, +39.1\%, and +35.4\% on NYUv2. 
%
%
Correspondingly, on KITTI (which features a depth range similar to TartanAir), the RMSE is reduced by $2.1 \times$, $1.5 \times$, and $1.5 \times$ for $48 \times 48$, $96 \times 96$, and $192 \times 192$ inputs, respectively. 
In contrast, on NYUv2, where the depth range differs substantially from TartanAir (0–5 m vs. 0–80 m), the RMSE improves by two orders of magnitude ($150 \times$, $131 \times$, and $108 \times$).

Despite these improvements, the relative accuracy gain decreases with increasing resolution. 
This effect is partially attributed to the upscaling of the $8 \times 8$ labels, which become less representative of the details of their respective images, while their resolution increases. 
Therefore, on KITTI, the maximum $\delta_1$ accuracy drops from 61.0\% at $48 \times 48$ to 52.8\% and 45.5\% at $96 \times 96$ and $192 \times 192$, respectively. 
On NYUv2, the corresponding values are 47.0\%, 43.7\%, and 39.9\%. 
Similarly, the RMSE after fine-tuning increases with resolution, reaching 9.4~m, 10.8~m, and 11.7~m on KITTI, and 0.5~m, 0.5~m, and 0.6~m on NYUv2. 

Increasing input resolution also affects the ODL memory cost. 
For $96 \times 96$ inputs, fine-tuning requires 3.5~MB ($2.9 \times$ more than $48 \times 48$), including 1.35~MB for the working buffer, while for $192 \times 192$, memory usage rises to 12.6~MB ($10.6 \times$), with 5.3~MB for the buffer. 
Overall, these results confirm that, even with larger input images, $8 \times 8$ labels can still provide effective in-field supervision. 


\subsection{Real-world Experiments} \label{sec:realdatasimulation}

\subsubsection{Accuracy on \texorpdfstring{IDSIA-$\mu$MDE}{IDSIA-μMDE} data} 

This section evaluates the performance of our $\mu$PyD-Net model in a real-world scenario, using the IDSIA-$\mu$MDE dataset. 
We test $\mu$PyD-Net, pre-trained on TartanAir, in two scenarios, i.e., without fine-tuning (baseline) and when fine-tuned with the memory-optimal sparse update setup selected in Sec.~\ref{sec:sparse_update_analysis}, which updates only DEC0.
We report the accuracy results of these configurations in Table~\ref{tab:idsiaccuracy}, comparing them for reference with the full update scheme and with a dummy predictor, which predicts the pixel-wise average depth of all testing samples of IDSIA-$\mu$MDE.

From an accuracy point of view, the baseline model poorly generalizes on IDSIA-$\mu$MDE. 
This can be noted by the RMSE (4.9~m), which exceeds the range of the on-board VL53L5CX depth sensor, used to collect labels in-field. 
Conversely, fine-tuning with both sparse and full update reduces the RMSE to 0.5~m in the former case, while in the latter to 0.6~m. 
However, the $\delta_1$ accuracy is very limited on the target (up to 30.8\% for the sparse update). 
This effect is caused by the noisy labels of IDSIA-$\mu$MDE: while the overall scale of the target environment can be learned from a few valid pixels, learning the pixel-wise depth is limited by the mismatch between the shapes of the objects in the images and their related depth. 

Furthermore, the sparse update scheme achieves better accuracy than a full update.
This effect can arise because only the early layers of the network are being updated. 
In this scenario, characterized by noisy labels, fine-tuning just the low-level features of the CNN helps improve the estimation of the coarse depth scale. 
Meanwhile, the prediction of higher-level details still benefits from the pre-training phase, during which the CNN was trained on highly accurate depth labels.

\begin{table}[t]
\centering
\caption{Accuracy evaluation on IDSIA-$\mu$MDE.}
\begin{tabular}{lccc}
\hline
\multicolumn{1}{c}{\textbf{Model}} & \textbf{$\delta_1$ $\uparrow$ {[}\%{]}} & \textbf{RMSE $\downarrow$ {[}m{]}} & \textbf{SiLog $\downarrow$} \\ \hline
Baseline \textit{(no fine-tuning)} & 17.0                                    & 4.9                                & 3.9                         \\
Sparse update \textit{(DEC0)}      & 30.8                                    & 0.5                                & 2.0                         \\ \hline
Full update                        & 27.4                                    & 0.6                                & 2.2                         \\
Dummy predictor                    & 7.1                                     & 2.0                                & 4.2                         \\ \hline
\end{tabular}
\label{tab:idsiaccuracy}
\end{table}

\begin{figure}[t]
\centering
\includegraphics[width=\linewidth]{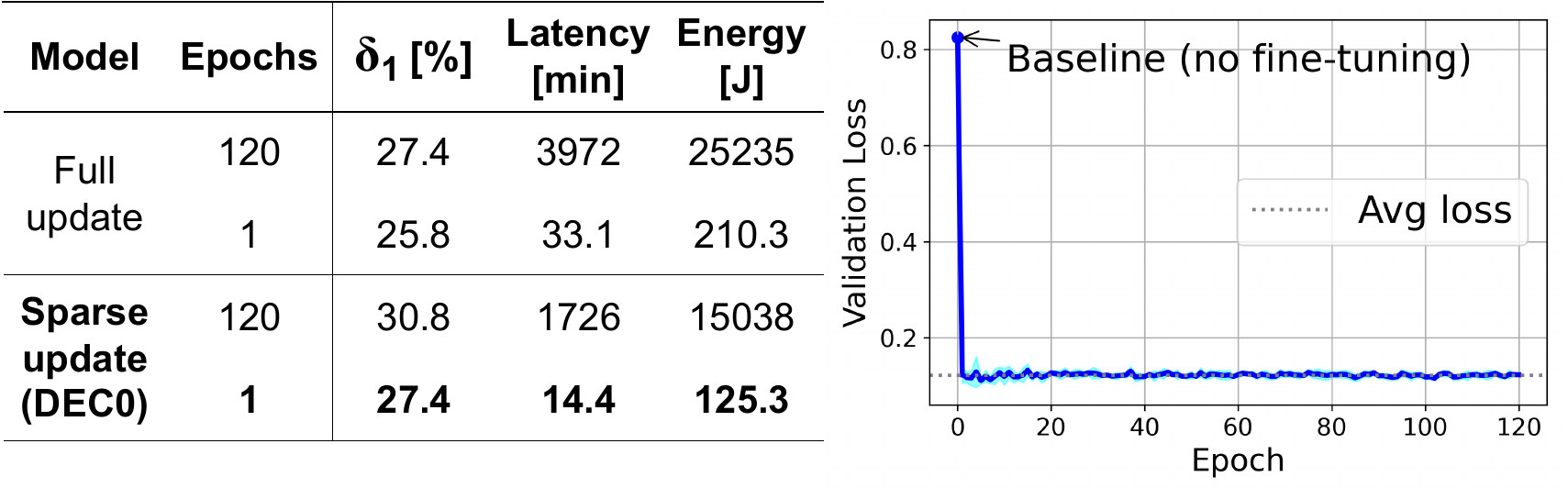}
\caption{Comparison between on-device sparse and full update varying the number of epochs and validation loss for sparse update on IDSIA-$\mu$MDE.}
\label{fig:idsiaumdeloss}
\end{figure}

\begin{figure}[t]
\centering
\includegraphics[width=\linewidth]{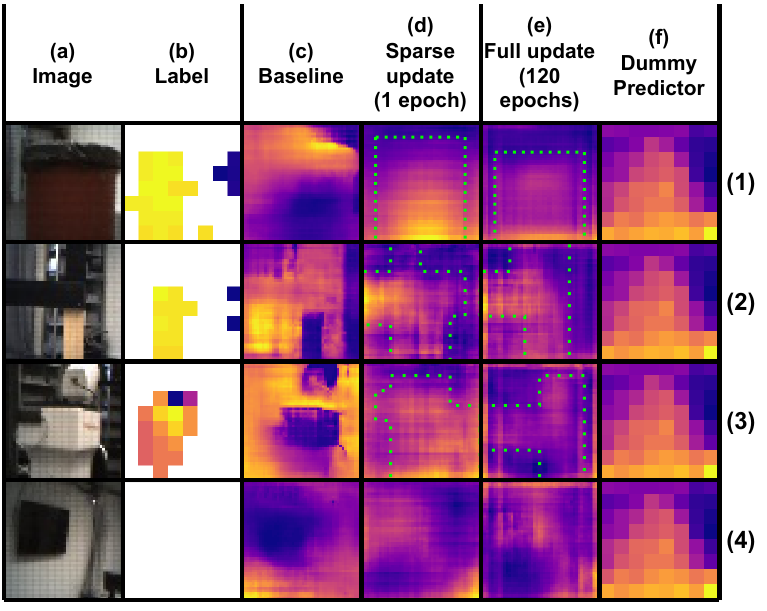}
\caption{Images, labels from IDSIA-$\mu$MDE and related $\mu$PyD-Net predictions.}
\label{fig:idsiaumdevisual}
\end{figure}

\subsubsection{ODL Measurements} 


Fig.~\ref{fig:idsiaumdeloss} reports the latency and energy measurements, on GAP9, of our on-device fine-tuning.
From a latency and energy point of view, fine-tuning $\mu$PyD-Net with sparse update requires 14.4~minutes on the GAP9 MCU per epoch, at the cost of 125.4~J.
In the case of a complete 120-epoch fine-tuning, these costs make training difficult to achieve on the MCU, requiring 66.2~hours and 25.3~kJ.
For our in-field setup, we can reduce such costs by minimizing the number of fine-tuning epochs performed on the MCU.
To this end, we study the evolution of the validation loss to observe when the model approximates convergence on the target dataset. 
Fig.~\ref{fig:idsiaumdeloss} reports this exploration over 120 epochs using sparse update. 
In this case, we note that $\mu$PyD-Net converges to the average loss value (0.122) in a few epochs ($<30$), already approximating such value ($0.123 \pm 0.02$) in the very first. 

Therefore, we can approximate the performance of a complete 120-epoch fine-tuning by limiting the training process to a single epoch.
Even with this limited setup, $\mu$PyD-Net achieves 0.6~m of RMSE and a $\delta_1$ accuracy of 27.4\%--3.4\% less than for 120 epochs--a result comparable to 120 epochs with full update.
Considering both phases of our ODL pipeline (Sec.~\ref{sec:method}), in which the acquisition of the 3035 training samples requires 3.37~minutes, our ODL routine can be completed in 17.8~minutes, thanks to the fine-tuning limited to a single epoch (requiring 14.4~minutes). 
Furthermore, this configuration significantly reduces the energy consumption of the MCU and the sensors to only 204.9~J for both data collection 79.5~J and fine-tuning (125.4~J). 

\subsubsection{Qualitative Results}
In Fig.~\ref{fig:idsiaumdevisual}, we report a qualitative analysis of our in-field results with four labeled images (Fig.~\ref{fig:idsiaumdevisual}-a,b) from IDSIA-$\mu$MDE.
As introduced in Sec.~\ref{sec:collect}, the labels of IDSIA-$\mu$MDE feature two sources of noise.
First, we notice how the mismatch between the FOVs of the camera and the depth sensors produces incorrect labeling on the observed objects (e.g., depth measurements on the right side of the depth maps of Fig.~\ref{fig:idsiaumdevisual}-1,2), leading to noisy training during the fine-tuning phase and a difficult pixel-wise evaluation during the testing phase. 
Second, we notice that the limited sensor range produces large regions with missing pixels (denoted in white), even leaving images (e.g., Fig.~\ref{fig:idsiaumdevisual}-4) with completely invalid labels. 

We show predictions of the baseline model (c) and the model fine-tuned for 1 epoch with sparse update (d), compared with the model fine-tuned for 120 epochs with full update (e) and the dummy predictor (f), highlighting detected objects with a green dotted line. 
Even for this dataset, the baseline model assigns a large depth to bright pixels.
However, for both images 1 and 4, it also detects the central part of the main objects (recycle bin, TV) as the farthest region of each image. 

In contrast, the fine-tuned models learn to distinguish the regions containing objects while imprecisely predicting details. 
In particular, (d) produces coarser predictions than (e), blurring the detected object and assigning its related depth to a wider area. 
Overall, we observe that the predicted depth is centered in the same position as the object seen in the image, rather than shifted, as in the depth labels collected in the field.
%
The dummy predictor, instead, only captures the correlation between the presence of obstacles in the central lower region of IDSIA-$\mu$MDE's images, which are surrounded by free space mostly on the upper right side. 

In conclusion, leveraging these fine-tuned models to estimate depth from images captured in the deployment environment provides notable advantages, even compared to the depth sensor used for generating the training labels.
Although learning from noisy and sparse labels, the models produce continuous domain-specific depth maps.
In contrast, the sensor is limited to detecting only nearby objects with sufficient reflectance and comes at a significant cost, being approximately $3\times$ more expensive than the camera itself.

Consequently, these fine-tuned MDE models offer a cost-effective alternative to the depth sensor, inferring depth from full scenes, including distant areas beyond the range of the sensor.
Moreover, the model predictions are not limited to surfaces with detectable reflectance. 
Instead, they provide depth estimates for entire objects—even in areas with low reflectance—though some uncertainty remains due to the inherent limitations of the pixel-level supervision during fine-tuning.
As a result, by producing dense and high-resolution depth maps from sparse sensor measurements, the system virtually expands the depth sensor’s resolution, allowing richer environmental understanding. 


\section{Discussion}\label{sec:discussion}

Our work presents the first ODL framework for MDE on IoT devices, addressing the challenge of autonomously determining labels for on-device adaptation. 
While label determination is a key step to fully exploit the potential of ODL, another critical aspect for achieving full autonomy is the detection of domain shift, i.e., identifying the conditions under which our ODL routine should be triggered.
In our multi-modal system, domain shift could be detected by periodically activating the on-board labeler—namely, the $8 \times 8$ depth sensor—to estimate the prediction error of the deployed model during inference. 
To validate this approach, we conducted preliminary experiments analyzing the $\delta_1$ accuracy distribution of our pre-trained model, computed using upscaled and simulated $8 \times 8$ labels, on TartanAir (in-domain data), and KITTI and NYUv2 (out-of-domain data).

Our analysis shows that the $\delta_1$ distributions form distinct clusters: in-domain data from TartanAir exhibit an average accuracy of $\delta_1 \approx 50.5\% \pm 21.9\%$, while out-of-domain samples from KITTI (centered at $\approx 14.7\%$) and NYUv2 ($\approx 2.3\%$) consistently fall below one standard deviation from the TartanAir mean. 
This clear separation indicates that a $\delta_1$-based in-field evaluation could effectively detect domain shifts. 
Consequently, the system could autonomously trigger ODL whenever the average $\delta_1$ accuracy, estimated in the field, drops below 28.6\%, a threshold that optimally distinguishes between in- and out-of-domain conditions.

Therefore, the estimation of the out-of-domain distribution can be performed during inference by activating the depth sensor at a low and constant rate (e.g., 0.1–0.5~Hz or lower), thereby ensuring uniform sampling of the data distribution while keeping the energy overhead minimal. 
These preliminary findings pave the way for future investigations on how to optimally tune the domain-shift detection threshold, ultimately enabling continuous learning in dynamic deployment environments.

\section{Conclusion}\label{sec:conclusion}

This work presented an ODL technique to address the fundamental issue of domain shift for MDE DNNs after deployment on MCU-powered IoT nodes.
Our method is based on a multi-modal low-resolution sensing frontend, composed of a monocular camera and  operates in two phases. 
First, the system collects noisy-labeled data from the new environment using the multi-modal sensors.
Then, the DNN model is fine-tuned on-device to be adapted to the new deployment environment. 
To enable on-device fine-tuning, we introduced a novel memory-driven sparse update scheme, targeting U-Net-like models, which reduces the memory requirements of backpropagation by $2.2 \times$, with respect to a full update. 
Thanks to the optimized deployment, based on the training kernels of PULP-TrainLib, our sparse ODL method can be performed in-field in 17.8~minutes, with an energy consumption of 204.9~J, making it possible to be executed on battery-operated nodes. 
To the best of our knowledge, our work is the first to target ODL for MDE on IoT nodes.
Future work will focus on studying the conditions to detect domain shift, as well as on the criteria for the selection of the most meaningful data for optimal fine-tuning.
To support the research effort on these topics, we release the code of our experiments, as well as the IDSIA-$\mu$MDE dataset. 

\section*{Acknowledgements}\label{sec:acks}

The authors thank Matteo Poggi for sharing technical details concerning $\mu$PyD-Net.

\bibliographystyle{unsrt} 
\bibliography{bibliography} 

\end{document}